\documentclass[journal]{IEEEtran}

\IEEEoverridecommandlockouts                              % This command is only needed if 
\usepackage[utf8]{inputenc}
\usepackage[T1]{fontenc}

\usepackage{graphicx}
\usepackage[namelimits]{amsmath} %数学公式
\usepackage{amssymb}             %数学公式
\usepackage{amsfonts}            %数学字体
\usepackage{mathrsfs}            %数学花体
\usepackage{caption}
\usepackage{subcaption}
\usepackage{booktabs}
\usepackage{cite}
\usepackage{url}
\usepackage{bbm}
\usepackage{hyperref}
\usepackage[switch]{lineno}
\usepackage{soul}
\usepackage{color}
\hypersetup{colorlinks=true, linkcolor=blue, filecolor=cyan, urlcolor=magenta}

\newcommand{\tabincell}[2]{\begin{tabular}{@{}#1@{}}#2\end{tabular}}

\begin{document}
\title{Efficient Deep Reinforcement Learning with Imitative Expert Priors for Autonomous Driving}

\author{Zhiyu Huang,
        Jingda Wu,
        Chen Lv,~\IEEEmembership{Senior Member,~IEEE}
        
\thanks{Z. Huang, J. Wu, and C. Lv are with the School of Mechanical and Aerospace Engineering, Nanyang Technological University, Singapore, 639798. (e-mails: zhiyu001@e.ntu.edu.sg, jingda001@e.ntu.edu.sg, lyuchen@ntu.edu.sg).}%
\thanks{This work was supported in part by the A*STAR Grant (No. 1922500046) and SUG-NAP Grant, Nanyang Technological University, Singapore.}%
\thanks{Corresponding author: C. Lv.}
}

\maketitle
\thispagestyle{empty}
\pagestyle{empty}

%%%%%%%%%%%%%%%%%%%%%%%%%%%%%%%%%%%%%%%%%%%%%%%%%%%%%%%%%%%%%%%%%%%%%%%%%%%%%%%%
\begin{abstract}
Deep reinforcement learning (DRL) is a promising way to achieve human-like autonomous driving. However, the low sample efficiency and difficulty of designing reward functions for DRL would hinder its applications in practice. In light of this, this paper proposes a novel framework to incorporate human prior knowledge in DRL, in order to improve the sample efficiency and save the effort of designing sophisticated reward functions. Our framework consists of three ingredients, namely expert demonstration, policy derivation, and reinforcement learning. In the expert demonstration step, a human expert demonstrates their execution of the task, and their behaviors are stored as state-action pairs. In the policy derivation step, the imitative expert policy is derived using behavioral cloning and uncertainty estimation relying on the demonstration data. In the reinforcement learning step, the imitative expert policy is utilized to guide the learning of the DRL agent by regularizing the KL divergence between the DRL agent's policy and the imitative expert policy. To validate the proposed method in autonomous driving applications, two simulated urban driving scenarios (unprotected left turn and roundabout) are designed. The strengths of our proposed method are manifested by the training results as our method can not only achieve the best performance but also significantly improve the sample efficiency in comparison with the baseline algorithms (particularly 60\% improvement compared to soft actor-critic). In testing conditions, the agent trained by our method obtains the highest success rate and shows diverse and human-like driving behaviors as demonstrated by the human expert. We also find that using the imitative expert policy trained with the ensemble method that estimates both policy and model uncertainties, as well as increasing the training sample size, can result in better training and testing performance, especially for more difficult tasks. As a result, the proposed method has shown its potential to facilitate the applications of DRL-enabled human-like autonomous driving systems in practice. The code and supplementary videos are also provided\footnote{\href{https://mczhi.github.io/Expert-Prior-RL/}{https://mczhi.github.io/Expert-Prior-RL/}}.
\end{abstract}


%%%%%%%%%%%%%%%%%%%%%%%%%%%%%%%%%%%%%%%%%%%%%%%%%%%%%%%%%%%%%%%%%%%%%%%%%%%%%%%%
%\linenumbers
\section{Introduction}
\IEEEPARstart{R}{einforcement} learning (RL) has seen successful applications in various domains including autonomous driving \cite{kiran2020deep} and intelligent transportation systems \cite{haydari2020deep}. Using an RL-based decision-making system can deliver a human-like driving experience (featuring interaction awareness and personalization), by learning to interact with other agents on the road without explicitly modeling the complex driving environment. However, there are two problems that RL needs to solve. One major problem of RL is the notoriously low data efficiency, which means an RL agent requires a massive amount of interactions with the environment to learn a functioning policy. It even gets worse as we attempt to tackle increasingly demanding and diverse problems in autonomous driving, such as crossing an unsignalized intersection or doing an unprotected left turn in dense traffic. Another serious challenge is the design of the reward function, which specifies the expected behaviors for the RL agent. A poorly designed reward function may cause the agent to mistakenly exploit the reward function and stick to unexpected behaviors, and thus it often takes a significant amount of time and effort to design a proper reward function and tweak the parameters. Although the reward functions can be learned from human driving data using some techniques like inverse reinforcement learning \cite{huang2020modeling, rosbach2019driving}, some structure of the reward function (e.g., a linear combination of different hand-crafted features) is usually assumed, which might not stand in practice. 

\begin{figure*}[htp]
    \centering
    \includegraphics[width=0.9\linewidth]{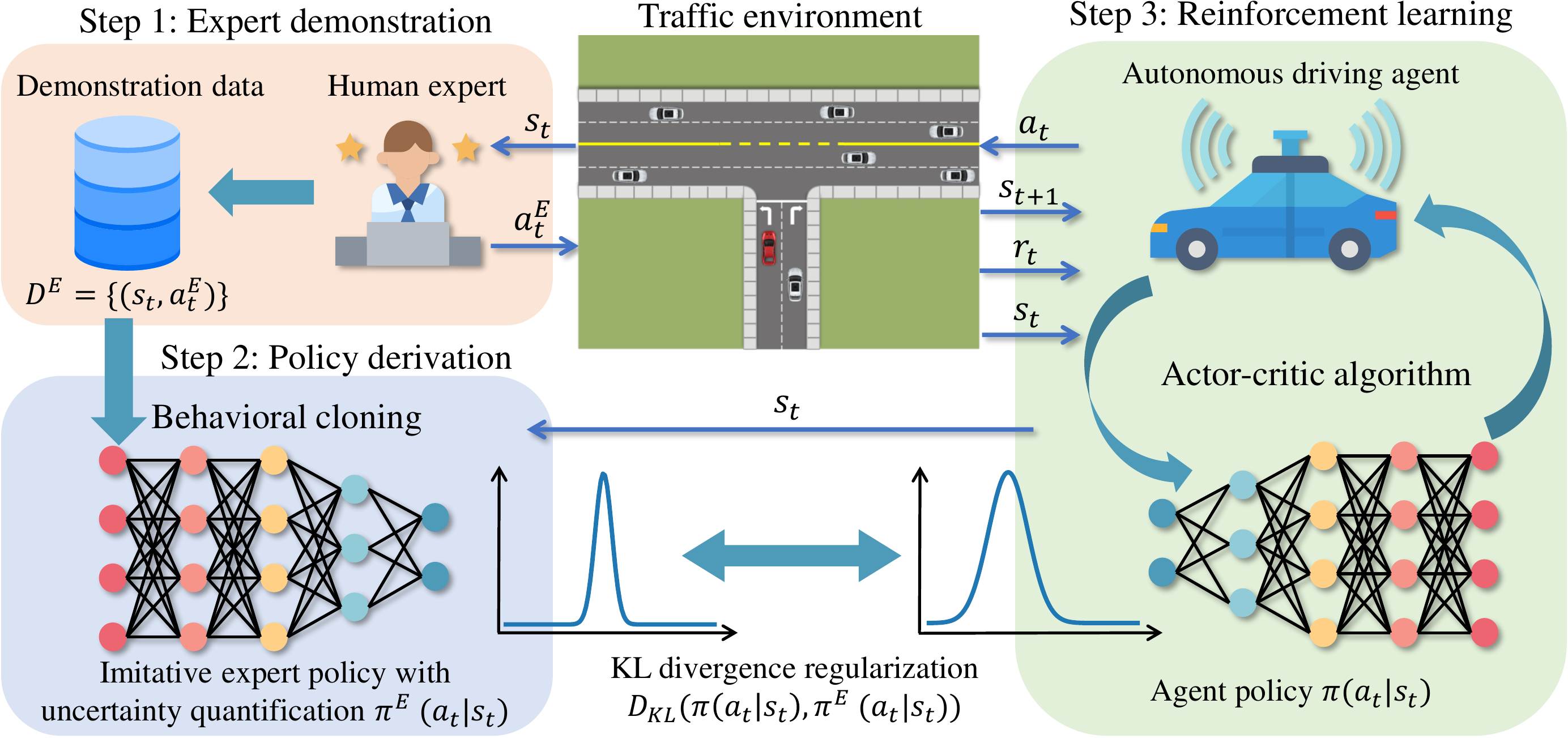}
    \caption{The conceptual framework of our deep reinforcement learning method with imitative expert priors}
    \label{fig1}
\end{figure*}

Facing these two major challenges, this paper proposes a novel RL framework that can transfer human prior knowledge into the RL agent with a small amount of human expert demonstration, in order to improve the sample efficiency and save the effort to design sophisticated reward functions. First of all, we distill the human prior knowledge through their demonstrations into the form of imitative expert policy using imitation learning and uncertainty estimation. Subsequently, the imitative expert policy is used to guide the learning process of RL agents through regularizing the RL policy to be close to the expert policy. The RL agent's exploration ability is still maintained if the uncertainty of the imitative expert policy is high. We also demonstrate that human-like driving behaviors are achievable with only a sparse reward function. In detail, as shown in Fig. \ref{fig1}, we first obtain a stochastic expert policy using imitation learning from expert demonstration. By adding the Kullback–Leibler (KL) divergence between the imitative expert priors and agent policy into the RL framework, we can regularize the agent's behaviors within the desired space, and thus the learning efficiency can be significantly enhanced. The main contributions of this paper are listed as follows. 

\begin{enumerate}
\item An RL framework that incorporates the imitative expert priors from expert demonstrations is proposed. The imitative expert priors are learned with imitation learning and uncertainty quantification method, which can regularize the RL agent's behavior while also maintaining its exploration ability.

\item Two different ways, namely value penalty and policy constraint, are proposed to integrate the imitative expert policy in the framework of the actor-critic algorithm and their performances are investigated. The superior sample efficiency of our proposed method is revealed in comparison with other RL algorithms.

\item The proposed method is intensively validated in challenging urban driving scenarios and we demonstrate that human-like driving behaviors emerge even the agent only gets sparse reward feedback from the environment. The effects of expert policies trained with different sample sizes and different uncertainty estimation methods are also investigated.
\end{enumerate}

%In the rest of this paper, Section \ref{sec2} introduces the background of the proposed method, particularly deep reinforcement learning and imitation learning with uncertainty estimation. Section \ref{sec3} elaborates the proposed RL algorithm with imitative expert priors regularization. In Section \ref{sec4}, the experimental setup and implementation details are stated, and the results and comparisons are presented in Section \ref{sec5}. The main conclusions are summarized in Section \ref{sec6}.

\section{Related Work}
\label{sec1.5}
\subsection{Imitation learning and reinforcement learning}
Imitation learning (IL) and reinforcement learning (RL) are two promising tools to be applied in autonomous driving, which can help develop adaptive, flexible, and human-like decision-making systems. Imitation learning, which employs an efficient supervised learning mechanism, has seen widespread use in autonomous driving research due to its simplicity and effectiveness. For example, imitation learning has been used in end-to-end autonomous driving systems that directly output control signals from raw sensor inputs \cite{hawke2020urban, huang2020multi} or modular driving systems that output planned trajectories from processed perception information \cite{Ogale-RSS-19}. However, one intractable drawback of imitation learning is the distribution shift, which means compounding errors lead to system failure due to the mismatch of training and inference distributions. On the other hand, RL will not encounter this problem because it relies on online interactions with the environment. Many works have utilized RL to tackle difficult tasks in autonomous driving, such as urban driving \cite{chen2021interpretable}, multi-agent interactions \cite{palanisamy2020multi}, and handling near-accident scenarios \cite{cao2020reinforcement}. However, the drawbacks of RL are obvious and hard to solve, which could impede its applications in practical use. For one thing, the training of RL is very inefficient, even in the simulated environment, and for another, the reward function is difficult to specify and tuning of the parameters is time-consuming.

\subsection{Learning with human prior knowledge}
To improve the sample efficiency of RL training or inject some desired properties (e.g., safety, smoothness, and risk-averse) into the RL-based policies, many works propose to include human prior knowledge in the RL framework. Human prior knowledge can be their understanding of the task or their guidance and demonstration provided to the RL agent, and it is revealed that leveraging human understanding together with RL methods can help solve autonomous driving tasks safely and efficiently. For example, \cite{chen2020autonomous, krasowski2020safe} proposed to add a safety check module encoding the expert-defined safety rules in the RL-based control system, to guide the exploration process and exclude dangerous actions, in order to prevent unsafe exploration and speed up training. \cite{duan2020hierarchical} relied on learning useful skills first (driving-in-lane, left lane change, and right lane change) and then a master policy to switch between different skills, to hierarchically and efficiently solve the driving task. In addition, utilizing human guidance or demonstration helps constrain the solution space and guide the RL agent's exploration, to eliminate unwanted interactions and accelerate the learning process. For example, \cite{wu2021human} proposed a real-time human-guidance-based learning method that lets human experts intervene in the training process in real-time and provide guidance and thereby enables the RL agent to learn from both human guidance and self-exploration. However, this kind of method is heavily dependent on online human guidance, which may impose a heavy workload on human experts. On the other hand, offline human demonstration is easy to obtain and does not require interactions between the RL agent and the human expert. Therefore, it is more favorable to incorporate human expert demonstration in the RL framework to enhance its performance.

\subsection{Reinforcement learning with demonstrations}
To make use of human demonstrations in RL, one approach is to pre-train the policy with imitation (supervised) learning using the expert demonstrations to initialize the policy with a reasonable level of performance, and then apply RL to obtain a better policy with respect to the reward function \cite{liang2018cirl, pfeiffer2018reinforced}. However, this method often does not work well, especially for the maximum entropy RL \cite{haarnoja2018soft2} that encourages randomness of the policy to explore at the beginning. An alternate approach is to add the expert demonstrations into the experience replay buffer for off-policy RL algorithms and sample the experience from both expert demonstrations and agent's interactions to update the policy \cite{vecerik2017leveraging, hester2018deep}. However, this kind of approach requires annotating reward for each state-action pair to comply with the format of transition tuples in RL, which is intractable in some cases due to the inaccessibility of the expert reward. \cite{wu2021human, liu2021improved} proposed to add imitation learning (minimizing the discrepancy between the agent actions and expert actions) to the policy update in addition to reinforcement learning (maximizing the Q values), to learn from both human demonstration and agent's self-exploration. However, a sophisticated design of reward function or reward shaping is still indispensable for these methods. On the other hand, our proposed method is inspired by the recent advances in offline RL \cite{levine2020offline, wu2019behavior}, which is to encourage the learned policy to be close to the behavior policy. This idea is originally brought forward to avoid value overestimation for actions outside of the training distribution, since the agent learns purely offline without online data collections. We propose to employ such a method in an online RL setting where the behavior policy is learned from expert demonstrations and then used to guide the exploration and constrain the RL agent's behaviors. To maintain the RL agent's exploration capability, we propose to use uncertainty estimation in deriving the expert policy with imitation learning, so that the agent can explore more when the uncertainty of the expert policy is high, meaning the action given by the imitative expert policy is not reliable. Moreover, the reward function design can be significantly simplified since the agent's behavior is regularized by the expert policy, thus allowing a sparse reward to work well even in complicated situations. 

\section{Background}
\label{sec2}
\subsection{Reinforcement learning}

Reinforcement learning (RL) confronts the problem of learning to control a dynamic system, which can be defined by a Markov decision process (MDP), represented as $ \mathcal{M} = \left( \mathcal{S}, \mathcal{A}, T, r, \gamma \right) $. In this tuple, $\mathcal{S}$ is the state space $\mathbf{s} \in \mathcal{S}$, $\mathcal{A}$ is the action space $\mathbf{a} \in \mathcal{A}$, $T$ is the transition probability distribution that defines the system dynamics $T(\mathbf{s}_{t+1}|\mathbf{s}_t, \mathbf{a}_t)$, $r$ is the reward function $r: \mathcal{S} \times \mathcal{A} \rightarrow \mathbb{R}$, and $\gamma \in (0, 1]$ is a discount factor. The target of RL is to acquire a policy, which is defined as a distribution over actions conditioned on states $\pi(\mathbf{a}_t|\mathbf{s}_t)$, in order to maximize the long-term discounted cumulative reward:
\begin{equation}
\label{eq1}
\max_{\pi} \ \underset{\tau \sim p_\pi(\tau)}{\mathbb{E}} \left[ \sum_{t=0}^T \gamma^t r(\mathbf{s}_t, \mathbf{a}_t) \right],
\end{equation}
where $\tau$ is a trajectory and $p_\pi(\tau)$ is the distribution of the trajectory under policy $\pi$, and $T$ is the time horizon.

One way to optimize the target is to estimate the state or state-action value function, and then recover a policy. The state value function (V-function) and state-action value function (Q-function) are recursively defined as:
\begin{equation}
V^{\pi} \left( \mathbf{s}_{t} \right) = \underset{\mathbf{a}_t \sim \pi{(\cdot|\mathbf{s}_t)}}{\mathbb{E}} \left[ Q^{\pi} \left( \mathbf{s}_{t}, \mathbf{a}_{t} \right) \right],
\end{equation}
\begin{equation}
Q^{\pi} \left( \mathbf{s}_{t}, \mathbf{a}_{t} \right) = r(\mathbf{s}_t, \mathbf{a}_t) + \gamma \mathbb{E}_{\mathbf{s}_{t+1} \sim T(\cdot|\mathbf{s}_{t}, \mathbf{a}_{t})} \left[ V^{\pi} (\mathbf{s}_{t+1}) \right].
\end{equation}

The optimal policy $\pi^*$ can be obtained by maximizing $V^{\pi^*}(\mathbf{s})$ at all states $\mathbf{s} \in \mathcal{S}$. In the actor-critic method, which is illustrated in details in Section \ref{sec3}, one optimizes a policy $\pi$ by alternating the learning of V-function and Q-function through minimizing the Bellman errors over one-step transitions and the learning of policy through maximizing the Q-values.

In addition to maximizing the cumulative discounted reward, soft actor-critic (SAC) \cite{haarnoja2018soft1, haarnoja2018soft2} employs a stochastic policy and adds an entropy term of the policy $\mathcal{H}$ to the target (Eq. (\ref{eq1})) so as to facilitate exploration:
\begin{equation}
\max_{\pi} \ \underset{\tau \sim p_\pi(\tau)}{\mathbb{E}} \left[ \sum_{t=0}^T \gamma^t \left( r \left( \mathbf{s}_t, \mathbf{a}_t \right) + \alpha \mathcal{H} \left( \pi \left( \cdot | \mathbf{s}_{t} \right) \right) \right) \right].
\end{equation}

The entropy term is equivalent to the KL divergence between the agent's policy and a uniformly random action prior. However, in our method, we regularize the learned policy $\pi$ towards the non-uniform and informative expert behavior policy $\pi^{E}$. In addition to that, another way is proposed, which is to directly constrain the discrepancy between the learned policy $\pi$ and the expert behavior policy $\pi^{E}$. Some methods of obtaining the expert policy $\pi^{E}$ from demonstration data are described below. 

\subsection{Behavioral cloning}
Since the expert policy is not directly accessible, one feasible and widely-used approach is to approximate the expert policy through imitation learning from expert demonstrations. Formally, given the expert demonstration dataset $\mathcal{D}^E: \{ \tau_i \}_N$, which consists of $N$ trajectories, and each trajectory is a sequence of state-action pairs $\tau = \{ \mathbf{s}_1, \mathbf{a}_1, \dots , \mathbf{s}_T, \mathbf{a}_T\}$, the imitative expert policy $\pi^{E}: \mathcal{S} \rightarrow \mathcal{A}$ can be obtained by maximizing the log-likelihood over $\mathcal{D}^E$:
\begin{equation}
\label{ml}
\pi^E := \underset{\pi}{\text{argmax}} \underset{(\mathbf{a}, \mathbf{s}) \sim \mathcal{D}^E}{\mathbb{E}} \left[ \log \pi(\mathbf{a}|\mathbf{s}) \right].
\end{equation}

In a general behavioral cloning setup, the expert policy is a neural network parameterized by $\theta$, which takes a vector of state inputs and generates a single deterministic output that represents the action, $\mathbf{a}_t = \pi_{\theta}(\mathbf{s}_t)$. Commonly, the policy network gets trained by minimizing the mean squared error (MSE) between the network outputs and expert actions, shown as:
\begin{equation}
\label{ml2}
\mathcal{L} (\theta) = \underset{(\mathbf{a}, \mathbf{s}) \sim \mathcal{D}^E}{\mathbb{E}} \parallel \pi_{\theta} \left( \mathbf{s} \right) - \mathbf{a} \parallel^2.
\end{equation}

Nonetheless, the above method can only yield a deterministic policy that outputs a point estimation of actions. This is not enough as a distribution of actions is needed to regularize the action distribution of the stochastic RL policy.

\subsection{Policy uncertainty}
In order to obtain a stochastic expert policy, we first consider the uncertainty of human (expert) behaviors, which means one can generate different feasible actions in the same state. We term this kind of uncertainty as policy uncertainty, which originates from the potential intrinsic randomness of human expert generating actions. Therefore, to account for this kind of uncertainty, we take the recourse of outputting the parameters of a parametric probability distribution over actions instead of deterministic actions, and we assume that the action is subject to a Gaussian distribution, i.e., $\mathbf{a}_t \sim \mathcal{N} \left( \mu_{\theta}(\mathbf{s}_t), \sigma^2_{\theta}(\mathbf{s}_t) \right)$. Subsequently, we employ the maximum likelihood approach to train the imitative expert policy, which is equivalent to minimize the negative log-likelihood (NLL) of the Gaussian distribution \cite{lakshminarayanan2017simple}:

\begin{equation}
\label{nll}
\mathcal{L} (\theta) = \underset{(\mathbf{a}, \mathbf{s}) \sim \mathcal{D}^E}{\mathbb{E}} \left[ \frac{\log \hat \sigma_{\theta}^2 (\mathbf{s}) }{2} + \frac{\left( \mathbf{a} - \hat \mu_{\theta}(\mathbf{s}) \right)^2}{2 \hat \sigma_{\theta}^2 (\mathbf{s})} + c \right],
\end{equation}
where $\theta$ is the parameters of the policy network, $\hat \mu_{\theta}$ and $\hat \sigma_{\theta}^2$ are the predictive mean and variance respectively, and $c$ is a constant.

\subsection{Model uncertainty}
\label{Model uncertainty}
Although we have derived a stochastic expert policy, the predictive mean and the variance of the policy are still uncertain and unreliable, for the data that are not included in the training dataset. We term this kind of uncertainty as model uncertainty, resulting from a lack of training data in certain areas of the state space, and it quantifies whether the model is confident about its action outputs. Estimating the model uncertainty is crucial for the imitative expert policy in our proposed method because the RL agent can often run into states that are not included in the demonstration dataset, thus requiring the confidence quantification on the out-of-distribution states. Some approaches have been put forward to estimate this kind of uncertainty, such as the Monte Carlo dropout method \cite{gal2016dropout, michelmore2020uncertainty} and deep ensemble method \cite{lakshminarayanan2017simple, tigas2019robust}. In this paper, we use the deep ensemble method because it is more computationally efficient and simple to implement. We adopt an ensemble of $M$ networks (stochastic policies) trained with different random initializations and orders of data using Eq. (\ref{nll}). We take all networks ($\theta_i$ referring to the parameters of $i$-th network) and combine their results into a Gaussian mixture distribution, and the mixture mean and variance can be calculated as:

\begin{equation}
\mu_{\pi^E}(\mathbf{s}) = \frac{1}{M} \sum_{i=1}^M \hat \mu_{\theta_i}(\mathbf{s}),
\end{equation}

\begin{equation}
\sigma_{\pi^E}^2(\mathbf{s}) = \frac{1}{M} \sum_{i=1}^M \hat \sigma^2_{\theta_i}(\mathbf{s}) + \left[ \frac{1}{M} \sum_{i=1}^M \hat \mu_{\theta_i}^2(\mathbf{s}) - \mu_{\pi^E}^2(\mathbf{s}) \right].
\end{equation}

We can treat the ensemble model as a Gaussian expert policy with mean $\mu_{\pi^E}(\mathbf{s})$ and variance $\sigma_{\pi^E}^2(\mathbf{s})$ that can capture both the policy uncertainty and model uncertainty.

%We can obtain the expert policy  $\pi^E(\mathbf{a}|\mathbf{s})$ , which is a Gaussian distribution over actions dependent on states 

\section{Deep reinforcement learning with imitative expert priors}
\label{sec3}

\subsection{Framework}
Fig. \ref{fig1} shows the conceptual framework of our method, which consists of three key steps: expert demonstration, policy derivation, and reinforcement learning. First and foremost, a human expert demonstrates their execution of the task, and their behaviors are encoded as sequences of state-action pairs. Then, the imitative expert policy is derived based on the demonstration data using methods from Section \ref{sec2}. The imitative expert policy assumes the action distribution the human expert would execute in the state, and it can be queried with a state encountered by the RL agent during training and responds with the reference action distribution. Finally, the imitative expert policy is applied to guide the learning process of the RL agent, by either adding a discrepancy penalty to the value function or regularizing the distributional discrepancy between the agent's policy and the expert one.

In the policy derivation step, we employ the stochastic policy setting and propose to estimate both the policy uncertainty and model uncertainty of the expert policy (as mentioned in Section \ref{Model uncertainty}), which is necessary and important. The intuition behind this is that when the uncertainty (variance) of the expert policy is small, meaning the policy is confident about its outputs, the RL agent's action should be close to the expert policy to avoid unnecessary exploration. In contrast, if the uncertainty (variance) is large, the mean value of the expert policy may not be sensible but it can admit a high variance, such that the resulting distribution could cover sensible actions. In this situation, the RL agent's action should also be close to the expert policy to explore more and find better actions. 

\subsection{Actor-critic algorithm}
For the agent learning step, we propose the modified actor-critic RL algorithm that can regularize the agent policy towards the imitative expert policy, and we present two ways to achieve that. The algorithm concurrently learns four networks, which are two Q-function networks ($Q_{\phi_1}$ and $Q_{\phi_2}$), a value function network ($V_\psi$), and a stochastic policy network ($\pi_\theta$). We separately keep a target value network ($V_{targ}$), which is obtained by polyak averaging the $V_\psi$ network parameters over the course of training. In addition, we introduce a pre-trained imitative expert policy network ($\pi^E)$, which can be an ensemble of networks or a single network, to generate the mean and standard deviation of the reference action distribution given a state input.

Both Q-function networks are learned with a single shared target, which is to minimize the mean-squared Bellman error (MSBE), shown as:
\begin{equation}
\label{qlearn}
\resizebox{0.96\linewidth}{!}{$
\mathcal{L}(\phi_i) = \underset{(\mathbf{s}_{t}, \mathbf{a}_{t}, r_t, \mathbf{s}_{t+1}) \sim \mathcal{D}}{\mathbb{E}} \left[ \left( Q_{\phi_i} \left( \mathbf{s}_{t}, \mathbf{a}_{t} \right) - \left( r_t + \gamma V_{targ} \left(\mathbf{s}_{t+1} \right) \right) \right)^{2} \right],$
}
\end{equation}
where $i=1,2$, $\mathcal{D}$ is the experience replay buffer consisting of a large number of transition tuples, and $V_{targ}$ is the target value function, which is updated once per the value network $V_{\psi}$ update by polyak averaging.

The value function network $V_\psi$ gets update by the following loss function:
\begin{equation}
\label{eq10}
\mathcal{L}(\psi) = \underset{ \substack{ {{\mathbf{s}}_t} \sim \mathcal{D} \\ {\widetilde {\mathbf{a}}_t} \sim {\pi _\theta }( \cdot |{{\mathbf{s}}_t})} }{\mathbb{E}} \left[ \left( V_{\psi} \left( \mathbf{s}_{t} \right) - \min_{i=1,2} Q_{\phi_i} \left( \mathbf{s}_{t}, \tilde{\mathbf{a}}_t \right) \right)^2 \right],
\end{equation}
where the actions are sampled fresh from the policy $\tilde{\mathbf{a}}_t \sim \pi_\theta \left( \cdot | \mathbf{s}_{t} \right)$, whereas the states should come from the replay buffer $\mathbf{s}_{t} \sim \mathcal{D}$. Only the minimum Q-value between the two Q-function networks is taken, to mitigate the overestimation in the value function.

The policy network $\pi_\theta (\cdot|\mathbf{s}_t)$ outputs the parameters of a Gaussian distribution, i.e., the mean $\mu_\theta$ and standard deviation $\sigma_\theta$, so that the final output action can be generated by sampling from this distribution, $\mathbf{a}_t \sim \mathcal{N} \left( \mu_\theta \left( \mathbf{s}_t \right), \sigma^2_\theta \left( \mathbf{s}_t \right) \right)$. The policy network is trained to maximize the value function at all states, which is equivalent to maximize the Q-value at any given state, and thus the loss function for the policy network becomes:
\begin{equation}
\label{eq11}
\mathcal{L}(\theta) = \underset{\substack{ {{\mathbf{s}}_t} \sim \mathcal{D} \\ {\widetilde {\mathbf{a}}_t} \sim {\pi _\theta }( \cdot |{{\mathbf{s}}_t})}}{\mathbb{E}} \left[ - \min_{i=1,2} Q_{\phi_i} \left( \mathbf{s}_{t}, \tilde{\mathbf{a}}_t \right) \right].
\end{equation}

Now that we have constructed the Q-learning and policy learning processes, we will introduce two approaches to incorporate the imitative expert priors into the actor-critic algorithm, in order to regularize the learned agent policy $\pi_\theta$ towards the expert policy $\pi^E$. Generally, there are two different ways: adding a divergence penalty to the value function (value penalty) or constraining the divergence between the agent policy and expert policy (policy constraint). The details of these two ways are given below.

\subsection{Expert priors as value penalty} 
To encourage the learned policy to be close to the expert policy, we can add a penalty term into the reward function, leading to the refactored reward function:
\begin{equation}
\bar r(\mathbf{s}_t, \mathbf{a}_t) = r(\mathbf{s}_t, \mathbf{a}_t) - \alpha D \left( \pi_\theta \left( \cdot | \mathbf{s}_t \right), \pi^E \left( \cdot | \mathbf{s}_t \right) \right), 
\end{equation}
where $\alpha$ is a temperature parameter, and $D$ is a probability metric to measure the divergence between two distributions over actions and is chosen to be the Kullback–Leibler (KL) divergence and denoted as $D_{KL}$ hereafter.

This is equivalent to directly incorporate the penalty term into the value function. Therefore, the updated loss function of the penalized value function can be rewritten as:
\begin{equation}
\begin{aligned}
\mathcal{L}(\psi) = \underset{ \substack{ \mathbf{s}_t \sim \mathcal{D} \\ \widetilde{\mathbf{a}}_t \sim \pi _\theta \left( \cdot | {{\mathbf{s}}_t} \right)} }{\mathbb{E}} \Bigg[ \bigg( V_\psi \left( \mathbf{s}_t \right) - \Big( \min_{i = 1,2}{Q_{\phi_i}} \left( \mathbf{s}_t, \widetilde{\mathbf{a}}_t \right) \\ 
- \alpha \hat D_{KL} \left(\pi_\theta \left( \cdot | \mathbf{s}_t \right), \pi^E \left( \cdot | \mathbf{s}_t \right) \right) \Big) \bigg)^2 \Bigg],
\end{aligned}
\end{equation}
where $\hat D_{KL}$ denotes the sample estimation of the KL divergence, $\pi_\theta \left( \cdot | \mathbf{s}_t \right)$ is the learned policy and $\pi^E \left( \cdot | \mathbf{s}_t \right)$ is the expert policy we have obtained through imitation learning. They are both modeled as the Gaussian distribution over actions conditioned on states.

Accordingly, the policy should maximize the penalized value function in each state, and thereby the loss function for the policy network should be modified as:
\begin{equation}
\begin{aligned}
\mathcal{L}(\theta) = \underset{\substack{ {{\mathbf{s}}_t} \sim \mathcal{D} \\ {\widetilde {\mathbf{a}}_t} \sim {\pi _\theta }( \cdot |{{\mathbf{s}}_t})}}{\mathbb{E}} 
\Bigg[ - \min_{i=1,2} Q_{\phi_i} \left( \mathbf{s}_{t}, \tilde{\mathbf{a}}_t \right) \\
+ \alpha \hat D_{KL} \left( \pi_\theta \left( \cdot | \mathbf{s}_t \right), \pi^E \left(\cdot | \mathbf{s}_t \right) \right) \Bigg].
\end{aligned}
\end{equation}

The update of the Q-function networks remains the same as Eq. (\ref{qlearn}), and now the actor-critic algorithm has been tailored to incorporate the behavior regularization through a penalty in the value function.

\subsection{Expert priors as policy constraint}
Instead of adding a penalty term to the value function, another way is to explicitly constrain the deviation between the learned policy and expert policy within a small value during policy optimization. Hence, the constrained optimization problem for policy learning can be formulated as:
\begin{equation}
\begin{aligned}
\min_\pi \underset{ \substack{ \mathbf{s}_t \sim \mathcal{D} \\ \widetilde{\mathbf{a}}_t \sim \pi _\theta \left( \cdot | {{\mathbf{s}}_t} \right)} }{\mathbb{E}} \left[ - \min_{i=1,2} Q_{\phi_i} \left( \mathbf{s}_{t}, \tilde{\mathbf{a}}_t \right) \right], \\
\text{s.t.} \ \hat D_{KL} \left( \pi_\theta \left( \cdot | \mathbf{s}_t \right), \pi^E \left(\cdot | \mathbf{s}_t \right) \right) \le \epsilon,
\end{aligned}
\end{equation}
where $\epsilon$ is a small positive constant or the tolerance of the KL divergence, which represents the closeness of the learned policy to the imitative expert policy.

To solve this constrained optimization problem, we can construct its corresponding Lagrangian dual problem:
\begin{equation}
\label{eq17}
\begin{aligned}
\max_{\lambda} \min_\pi \underset{ \substack{ \mathbf{s}_t \sim \mathcal{D} \\ \widetilde{\mathbf{a}}_t \sim \pi _\theta \left( \cdot | {{\mathbf{s}}_t} \right)} }{\mathbb{E}} \Bigg[ - \min_{i=1,2} Q_{\phi_i} \left( \mathbf{s}_{t}, \tilde{\mathbf{a}}_t \right)\\
+ \lambda \left(\hat D_{KL} \left( \pi_\theta \left( \cdot | \mathbf{s}_t \right), \pi^E \left( \cdot | \mathbf{s}_t \right) \right) - \epsilon \right) \Bigg],
\end{aligned}
\end{equation}
where $\lambda$ is the non-negative Lagrangian multiplier $\lambda \ge 0$.

We can employ the dual gradient descent method to optimize the unconstrained objective by alternating gradient descent steps on $\theta$ and $\lambda$ respectively \cite{siegel2020keep}. The core idea is to first solve the Lagrangian function (Eq. (\ref{eq17})) with respect to the policy $\pi_\theta$ under the current Lagrangian multiplier $\lambda$ (starting with a random guess), and then update the Lagrange multiplier $\lambda$ if the constraint is violated. Therefore, the modified loss function for the policy network is:
\begin{equation}
\label{eq16}
\begin{aligned}
\mathcal{L}(\theta) = \underset{\mathbf{s}_t \sim \mathcal{D}}{\mathbb{E}} 
\Bigg[ \underset{\widetilde{\mathbf{a}}_t \sim \pi_\theta \left( \cdot | {{\mathbf{s}}_t} \right)}{\mathbb{E}} - \min_{i=1,2} Q_{\phi_i} \left( \mathbf{s}_{t}, \tilde{\mathbf{a}}_t \right) \\
+ \lambda \left(\hat D_{KL} \left( \pi_\theta \left( \cdot | \mathbf{s}_t \right), \pi^E \left( \cdot | \mathbf{s}_t \right) \right) - \epsilon \right) \Bigg].
\end{aligned}    
\end{equation}

The Lagrangian multiplier $\lambda$ gets updated according to the following loss function with the constraint $\lambda \ge 0$:
\begin{equation}
\mathcal{L}(\lambda) = \underset{\mathbf{s}_t \sim \mathcal{D}}{\mathbb{E}} \left[ - \lambda \left( \hat D_{KL} \left( \pi_\theta \left( \cdot | \mathbf{s}_t \right), \pi^E \left( \cdot | \mathbf{s}_t \right) \right) - \epsilon \right) \right].
\end{equation}

For Q-learning, the update of the value function network remains the same as Eq. (\ref{eq10}), and the update of the Q-function networks is consistent with Eq. (\ref{qlearn}).

\section{Experimental Setup}
\label{sec4}

\subsection{Driving scenarios}
\begin{figure}[htp]
    \centering
    \includegraphics[width=\linewidth]{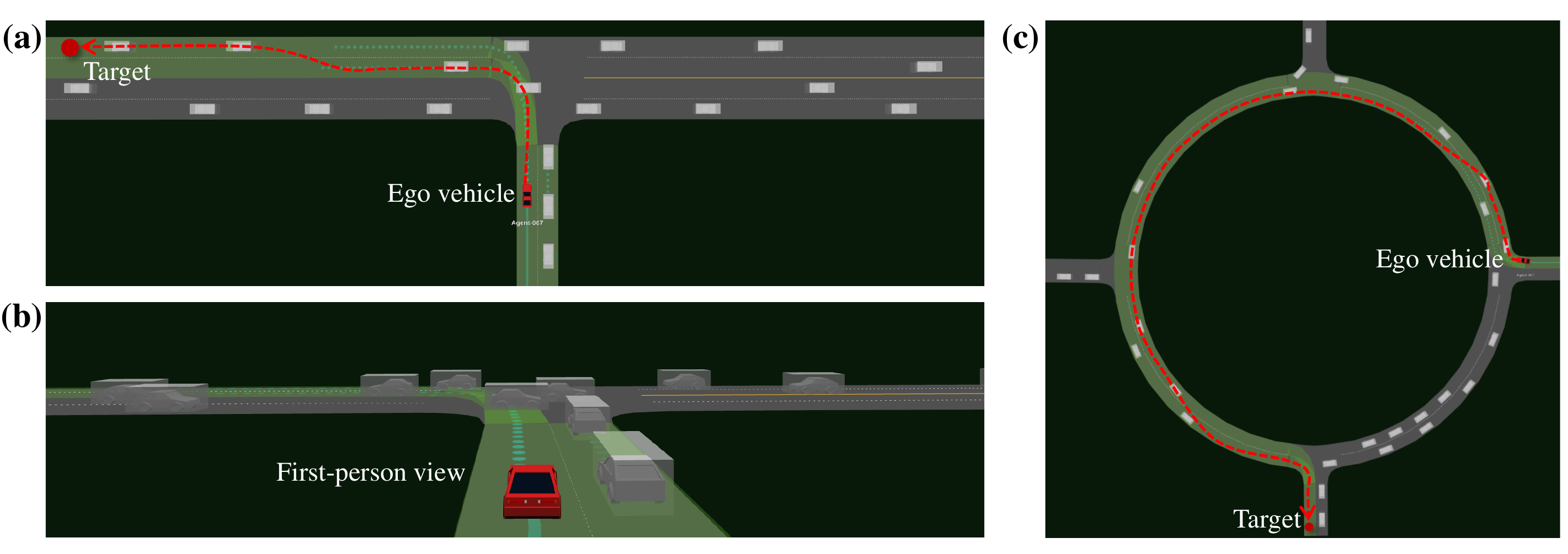}
    \caption{The designed urban driving scenarios in the SMARTS platform: (a) unprotected left turn; (b) first-person view for expert demonstration; (c) roundabout.}
    \label{fig2}
\end{figure}

We validate our proposed algorithm in solving urban autonomous driving problems. Two challenging urban driving scenarios are devised based on the SMARTS simulation platform \cite{zhou2020smarts}. In Fig. \ref{fig2}(a), the autonomous driving agent is required to complete an unprotected left turn task in heavy traffic. The agent needs to turn left onto a major road, crossing a two-way four-lane road and reaching the rightmost lane, without the regulation of traffic lights. Fig. \ref{fig2}(b) shows a first-person view from the ego vehicle in the left turn scenario, which is for a human expert to observe the environment. In Fig. \ref{fig2}(c), a roundabout scenario is designed and the autonomous driving vehicle needs to safely navigate from the start point to the destination. The green area in Fig. \ref{fig2} shows the accessible routes of the ego vehicle and the waypoints ahead of the ego vehicle (up to 50 meters) are displayed. These two scenarios are sufficiently different and challenging to validate the capabilities of our proposed method from different angles. The left turn scenario is more complex in terms of the traffic situations yet the vehicle maneuver is easier (focusing on speed control) and it requires a shorter time to finish; the traffic in the roundabout scenario is less complex, but it requires a longer time to complete and involves more subtle maneuvers in addition to avoiding collisions, such as changing lanes to avoid congestion.

Since there could be different types of drivers on the road, we need to generate different kinds of driver behaviors for the environment vehicles, which can help simulate real-world situations. Particularly, we randomly choose the parameters of the driving behavior models (e.g., speed distribution, maneuver imperfection, the impatience of waiting at intersection, willingness to cooperate) for a specific type of actor in a reasonable range, and each traffic flow consists of a mix of different types of actors. We generated 20 different traffic flows for training, and the initial conditions (e.g., vehicle spawn points) are randomly generated by the simulator at the beginning of each episode. In testing, we generate another 50 traffic flows to test the generalization capability of our trained policy.

\subsection{MDP formulation}
\subsubsection{State space} 
we use a bird-eye view image encoding the environment information within a square area of $32 \times 32$ meters with the ego vehicle at the center, which is assumed to be obtained from an upstream perception module. The bird-eye view image is an RGB image of the abstracted driving scene (see Fig. \ref{fig3}), in which the grey area is the road, the red rectangle is the ego vehicle, and the white rectangles are the surrounding vehicles. Since such a representation encodes a rich amount of information about the road structure and interaction situations and can cope with an arbitrary number of objects, it is suitable for highly interactive urban driving scenarios. The size of the bird-eye view image is $80 \times 80 \times 3$, which means the resolution of the detection area is 0.4 m/pixel. We use a stack of three consecutive bird-eye-view images of the current time step and two previous time steps to include a history of observations, and therefore, the shape of the state space is $80 \times 80 \times 9$.

\subsubsection{Action space}
we adopt a lane following controller that allows a high-level target input (i.e., target speed $V_T$ and lane change $L_T$) to control the vehicle rather than direct vehicle control commands (e.g., steering and accelerator). The longitudinal motion control is the continuous target speed $V_T$ within the range of $[0, 10]$ m/s, which is then normalized into $[-1, 1]$. The lateral motion is determined by the discrete lane change action $L_T$, namely change to the left lane $L_T=-1$, lane-keeping $L_T=0$, and change to the right lane $L_T=+1$. To adapt the discrete action into a continuous value for the lane-change control, we discretize a continuous range $[-1, 1]$ into three equal-sized bins, and each bin represents a discrete action, e.g., $[-1, -1/3]$ belongs to the action of change to the left lane and $[-1/3, 1/3]$ belongs to the lane-keeping action. The lane-changing action executed at a decision-making step will not finish the whole lane-change maneuver, and therefore, to finish a lane-change maneuver, the policy must output the lane-change action during consecutive decision-making steps. The reason why we choose this control setting is that controlling the longitudinal motion requires more refined and precise actions, while for the lateral motion control, discrete lane following or lane changing actions would be enough as the vehicle follows the road structure and traffic rules and keeps the lane most of the time.

\subsubsection{Reward function}
we use a sparse reward function, which only emits a meaningful value at the end of a training episode: 
\begin{equation}
r(s, a) = r_{\text{collision}} + r_{\text{goal}},
\end{equation}
where $r_{\text{collision}}=-1$ is the penalty for colliding with other objects and $r_{\text{goal}}=1$ is the reward for reaching the goal position, which are all zeros otherwise.

However, we find out that the sparse reward makes it very difficult for the baseline RL algorithms to achieve at least an acceptable performance. Therefore, to improve the performance of the baseline RL algorithms for a fair comparison, we employ the reward shaping technique and the shaped reward function $r^\prime (s, a)$ is:
\begin{equation}
r^\prime (s, a) = 0.001r_{\text{speed}} + r_{\text{collision}} + r_{\text{goal}},
\end{equation}
where $r_{\text{speed}}$ is the speed of the vehicle. Note that this shaped reward function is only used in other RL algorithms to improve their performance, and to prevent the disturbance of the reward-shaping term to our proposed method.

\subsection{Expert demonstration}
We ask a human expert to execute the designed urban driving tasks in the simulator and collect their actions as expert demonstration data. A human expert observes the driving environment in a first-person view (see Fig. \ref{fig2}(b)) and provides their actions, and there are four discrete actions that a human expert can manipulate through a keyboard. For the longitudinal direction, two discrete actions, speed up (increase the speed by 2 m/s) and slow down (decrease the speed by 2 m/s), are used to set the target speed for the vehicle. For the lateral movement, two discrete actions (change to the left lane and the right lane) are used, and the default maneuver is to keep the current lane. The human expert can execute the longitudinal and lateral control actions simultaneously but need not provide actions at every timestep. The vehicle will adhere to executing the last set target until the target has been altered by the expert. We also ask the expert to demonstrate different behaviors in the unprotected left turn task, e.g., aggressive behavior like nudging forward in the intersection and conservative behavior like stopping to find an acceptable gap. The demonstration data is stored in the format of state-action pairs and only the successful executions of the task count. The states are normalized to $[0, 1]$ and the actions are normalized to $[-1, 1]$. We collect 40 trajectories of human expert demonstration in the roundabout scenario, and for the left turn scenario, we collect 40 trajectories each for two distinct driving behaviors. We also investigate the effects of sample size in training imitative expert policy on the training and testing results of the RL agent.

\subsection{Comparison baselines}
Some RL and IL methods are implemented as the baselines to benchmark the performance and efficiency of the proposed method. The baselines methods are listed as follows.

\begin{itemize}
\item Soft actor critic (SAC) \cite{haarnoja2018soft1}. SAC is a state-of-the-art off-policy RL algorithm, which optimizes a trade-off between the expected return and entropy. We adopt the version that can automatically tune the entropy parameter. 

\item Proximal policy optimization (PPO) \cite{schulman2017proximal}. PPO is a state-of-the-art on-policy RL algorithm that can stabilize the policy training by limiting the new policy not getting far from the old policy using a clipped surrogate objective function.

\item Generative adversarial imitation learning (GAIL) \cite{ho2016generative}. GAIL uses expert demonstration to recover a policy using generative adversarial training. The policy is the generator that gets trained with an RL algorithm using a surrogate reward given by the discriminator, which measures the similarity between generated and demonstration samples.

\item Behavioral cloning (BC). BC is a supervised learning method with the objective of reproducing the expert demonstration actions at given states (see Section \ref{sec2}). We train a stochastic policy using the behavioral cloning method with uncertainty estimation for comparison, and for convenience, we directly use the imitative expert policy as the BC baseline. Since the learning mechanism of BC is different from RL training, it is only compared against our method in the testing phase.
\end{itemize}

\subsection{Implementation details}
\begin{figure}[htp]
    \centering
    \includegraphics[width=\linewidth]{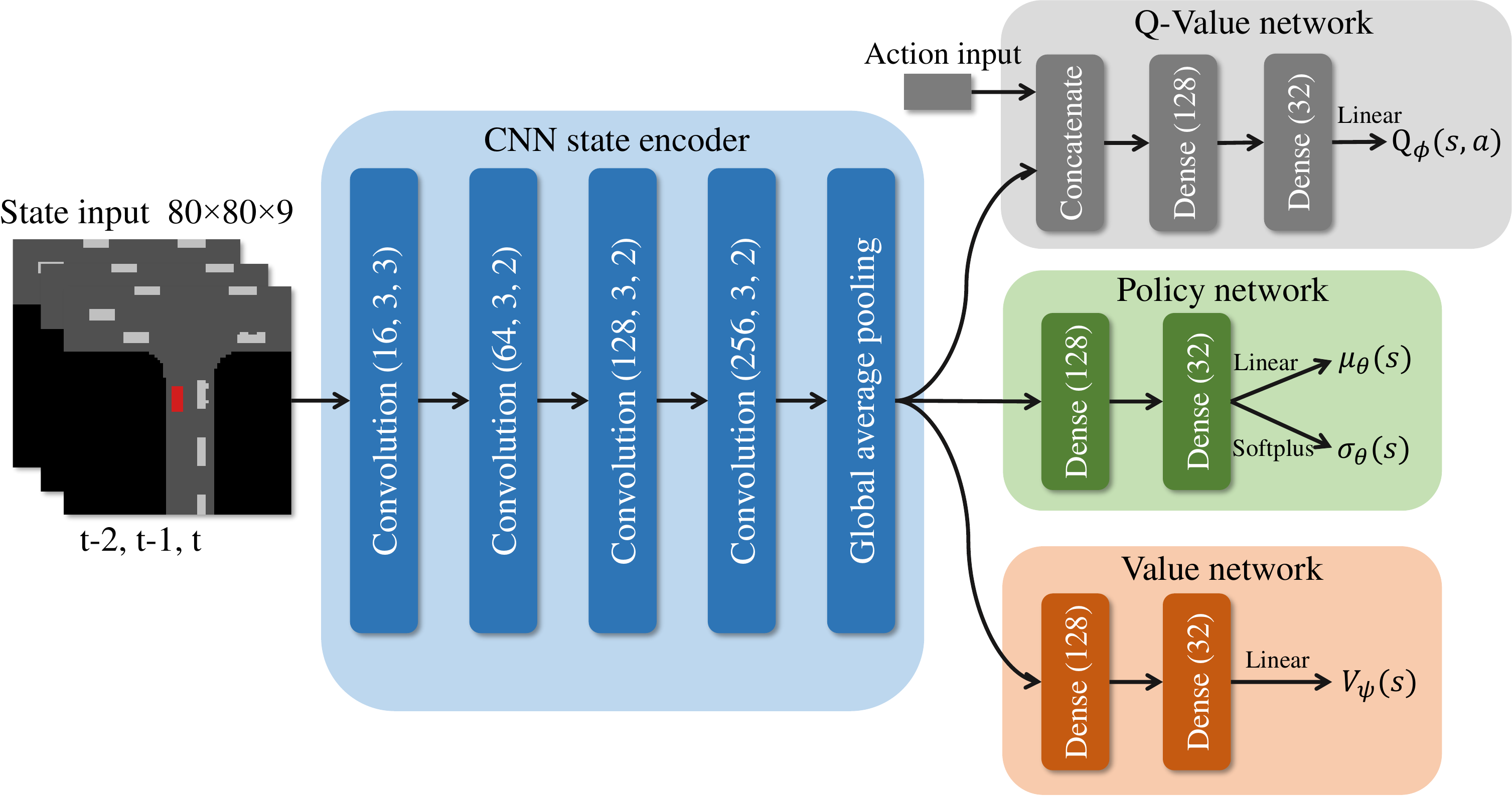}
    \caption{The structures of the neural networks}
    \label{fig3}
\end{figure}

The structure of the neural networks are shown in Fig. \ref{fig3}. All the networks share the same structure of state encoder, which consists of four convolutional layers and a global average pooling layer, followed by different structures of fully connected layers for different networks to generate desired outputs. The hyperparameters of the convolutional layers are given in the figure in a tuple format, following the meaning of (number of convolution kernels, kernel size, and moving stride). The numbers of hidden units of the fully connected layers are also given. The action distribution, which consists of two continuous actions with respect to longitudinal and lateral motion control, is modeled as a multivariate Gaussian distribution. The policy network generates the parameters of the Gaussian distribution, and the value and Q-value networks output a single value, respectively. All the layers, if not explicitly notated in Fig. \ref{fig3}, use the ReLU activation function. All the networks are trained with Adam optimizer in Tensorflow using an NVIDIA RTX 2080 Super GPU. 

For training the ensemble expert policy model, we train five single policy networks with different random seeds, both in the initialization of the weights of the network and shuffle of the dataset, and each network in the ensemble is trained for 100 epochs. The policy network in the ensemble employs the same structure as the policy network in Fig. \ref{fig3}. Since the expert demonstration actions (target speed, lane selection) are recorded in discrete values, directly regressing on these discrete values makes the network hard and unstable to train and performs very badly. Therefore, we add a small normally distributed random number to the discrete actions, which would not change the actual action very much, yet make the neural network to be easy and stable to train. The standard deviation of the Gaussian distribution of the obtained expert policy model is added by 0.1 to form a final output action distribution, which is to ensure that the distribution is wide enough to cover feasible actions.

The simulation runs on an AMD Ryzen 3900X CPU and takes roughly one hour to finish 100,000 steps of interaction with the simulated environment. The simulation interval is 0.1 seconds, which means the RL agent makes decisions every 0.1 seconds. The hyperparameters used in this paper are carefully tuned to achieve the best training performance and the final settings are listed in Table \ref{tab1}.

\begin{table}[htp]
\caption{Hyperparameters used in the experiment}
\label{tab1}
\centering
\begin{tabular}{@{}ccc@{}} 
\toprule
Symbol              & Meaning                       & Value \\ \midrule
$M$                 & Number of the ensemble models & 5     \\
$\alpha$            & KL divergence parameter       & 0.002    \\
$\lambda_0$         & Initial Lagrange multiplier   & 0.01     \\
$\epsilon$          & KL divergence tolerance       & 0.8    \\
$N_B$               & Size of the replay buffer     & 20000      \\
$\mathcal{B}$       & Training batch size           & 32      \\
$lr$                & Learning rate                 & 0.0003     \\
$\gamma$            & Discount rate                 & 0.99        \\
$N_{\text{warm}}$   & Warm-up steps                 & 5000      \\
$N_{\text{train}}$  & Total training steps          & 100000     \\ \bottomrule
\end{tabular}
\end{table}

\section{Results and Discussions}
\label{sec5}

\subsection{Training results}

We apply the proposed method along with baseline methods to train an autonomous driving agent in the designed driving scenarios. In the unprotected left turn scenario, the human expert purposely demonstrates two distinct driving styles, namely conservative and aggressive. Conservative behavior is reflected by that a driver would always stop at the intersection, waiting for enough gap to cross without interrupting other vehicles, whereas aggressive behavior means that a driver would nudge forward in the intersection to show their aggressiveness, forcing other vehicles on the main road to yield. In the roundabout scenario, the goal of the human expert is to pass through the roundabout as quickly as possible and avoid collisions, to ensure both efficiency and safety. The imitative expert policy here is derived with policy and model uncertainty estimations (ensemble model), and the effects of uncertainty estimation will be discussed subsequently. For each method, ten trials are conducted with different random seeds to reflect the average training performance of that method. Because the proposed method and baseline methods use different reward functions, the scales of the average episodic reward are different, and thereby we use the average success rate to gauge the training performance. It is defined as the number of successfully finished episodes over the last 20 episodes divided by 20. Success is counted only when the autonomous driving vehicle reaches the goal in a limited time. The failure conditions include colliding with other vehicles, going off the drivable road, and exceeding the maximum allowed time. The maximum allowed time for each episode in the left turn scenario is 40 seconds and 60 seconds for the roundabout scenario.
 
\begin{figure*}[htp]
    \centering
    \includegraphics[width=\linewidth]{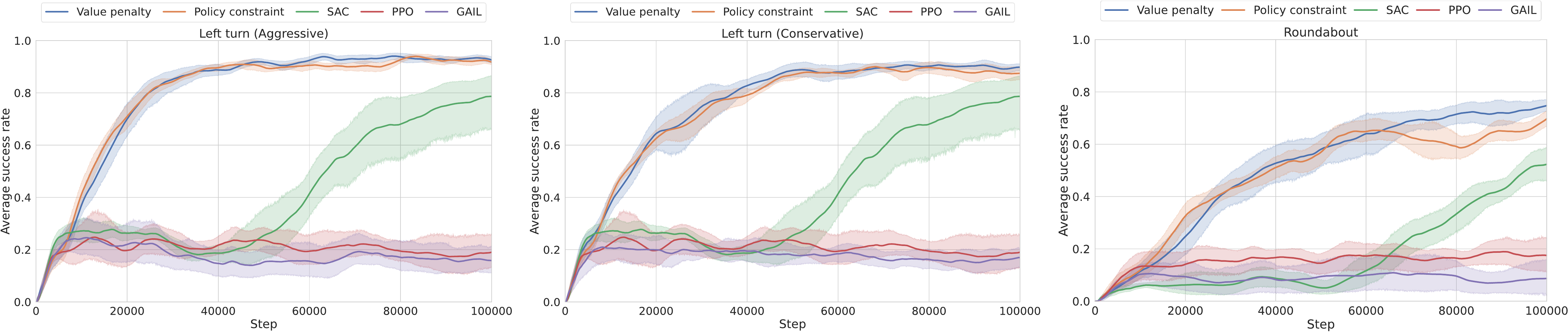}
    \caption{The training processes of different learning methods in the urban driving scenarios: the left shows the average success rate in the unprotected left turn scenario with aggressive expert policy, the middle shows that of the left turn scenario with conservative expert policy, and the right shows that of the roundabout scenario.}
    \label{fig4}
\end{figure*}

Fig. \ref{fig4} shows the training results of different methods in the urban driving scenarios, in which the training curve is represented by a solid line of the mean value and an error band of the 95\% confidence interval. To smooth the training curve, the average success rate at each step is the exponentially weighted moving average over the immediate 3000 steps. In the unprotected left turn scenario, our proposed methods have achieved the best performance at the end of the training process, and the value penalty method and policy constraint method perform equally well in this scenario. In addition to the boost in training performance, our method shows a significant improvement in sample efficiency. It only takes about 30\% of interaction steps to reach the same performance as that of SAC, improving the sample efficiency by 70\%. This is because while SAC is randomly searching for better actions, the imitative expert policy used in our method could provide a reasonable direction for searching for desired actions, thus remarkably enhancing the sample efficiency. Besides, the training process of SAC is unstable across different trials, which is reflected in the wide error band (high deviation). In contrast, the superior performance of our proposed method is consistent across different trials. The other two baseline methods, PPO and GAIL, do not perform very well. Their performance is hardly improved during the training process and the training curve almost remains flat. This is because for PPO, it can easily get trapped in local optima, and for GAIL, the adversarial training setting is hard and parameter-sensitive, which may cause it to perform poorly with high-dimensional state input. In terms of the difference in learning from different driving behaviors, we can tell that learning conservative behavior is harder than learning aggressive behavior. This is probably because stopping to find enough gaps is a very subtle behavior and hard to learn, and there is ambiguity about when to start crossing. Moreover, we can find out aggressive driving behavior can lead to a higher success rate. One possible reason is that the environment vehicles are intended to yield to the ego vehicle even if it violates the safety requirement, and aggressive driving behavior may exploit this feature.

In the roundabout scenario, our proposed method still demonstrates its advantages, achieving a higher success rate and better sample efficiency than the other methods. Unlike the left turn scenario where decisions are made in a small fragment of the scene (the intersection), the roundabout scenario is larger, involving more complicated situations, and requires a longer time to finish the task. Therefore, the value penalty method is more favorable than the policy constraint method in this situation. This is because the reward feedback from the environment is sparse, and adding additional feedback to the reward could help better estimate the value of actions and states. Compared to SAC, our proposed method sees a 60\% improvement in sample efficiency and a better success rate at the end of training. PPO and GAIL still act very poorly, and they can hardly improve the policy during the training. 

\subsection{Testing results}
We now begin to test the performance of the learned driving policies. We first generate 50 traffic flows that are different from the training situations and then let the driving policy with the best training return control the ego vehicle to navigate from the start point to the target point. In testing, we only take the mean value of the stochastic policy (Gaussian distribution policy) as the output action instead of sampling from it. All the learned driving policies obtained from different learning methods are tested for 50 episodes, and the testing success rate in percentage, defined as the number of success episodes divided by the number of testing episodes, is used to measure the testing performance or generalization capability. The results are given in Table \ref{tab2}. 

\begin{table}[htp]
\caption{Testing success rate (\%) of different learning methods in the urban driving scenarios (A: aggressive, C: conservative)}
\label{tab2}
\centering
%\resizebox{\linewidth}{!}{%
\begin{tabular}{@{}cccc@{}}
\toprule
                                & \tabincell{c}{Unprotected \\ left turn (A)} & \tabincell{c}{Unprotected \\ left turn (C)}  & Roundabout    \\ \midrule
BC                              &  72                                         & 70                                           &  36          \\
GAIL                            &  26                                         & 24                                           &  32           \\
PPO                             &  64                                         & 64                                           &  34           \\
SAC                             &  78                                         & 78                                           &  74           \\
Ours (policy constraint)        &  92                                         & 90                                           &  80           \\ 
Ours (value penalty)           &  \textbf{96}                                & \textbf{96}                                  &  \textbf{84}  \\\bottomrule
\end{tabular}
%}
\end{table}

According to the data in Table \ref{tab2}, we can conclude that the testing performance of each method is generally in line with their training performance. Our proposed value penalty method achieves the highest testing success rate in all three situations with different driving scenarios and demonstration behaviors. The simple behavioral cloning method can outperform GAIL and PPO, partially due to the low sample efficiency of on-policy RL algorithms. Nonetheless, our proposed method can achieve much better performance with the same number of training steps, which demonstrates higher sample efficiency. SAC performs quite well in both unprotected left turn scenario and roundabout scenario in terms of success rate but fails to compete against our proposed method. In addition, we list the duration of the success episodes for each learning method in Table \ref{tab3}, which clearly shows the different outcomes of driving behaviors when dealing with the unprotected left turn scenario. The duration of the driving policies with conservative driving behavior is four to five seconds longer than that of aggressive behavior, because the vehicle needs to wait for enough gap in the intersection. For the SAC agent, the duration is significantly shorter than other methods, because it is motivated to reach the target as quickly as possible with a higher speed. However, we have to note that the behavior of SAC is much more aggressive (rushing through the intersection), and thus safety is sacrificed to some extent and more collisions happen in the intersection. In contrast, our method can learn from a human expert that shows a balance between safety and efficiency, as well as different driving styles. In the roundabout scenario, the duration of GAIL and PPO agents is much longer than SAC and our proposed method because the vehicle only runs at a slower speed. Although the SAC agent can run faster than GAIL and PPO, it often sticks on the outer lane and does not learn to change lanes to gain a higher speed, and thus requires a longer time to finish compared to our proposed method. On the contrary, our proposed method with human demonstration can learn to change lanes to run faster and complete the task faster. 

\begin{table}[htp]
\caption{Testing duration (seconds) of different learning methods in the urban driving scenarios (A: aggressive, C: conservative)}
\label{tab3}
\centering
\resizebox{\linewidth}{!}{%
\begin{tabular}{@{}cccc@{}}
\toprule
                                & \tabincell{c}{Unprotected \\ left turn (A)} & \tabincell{c}{Unprotected \\ left turn (C)}  & Roundabout    \\ \midrule
BC                              &  $14.89\pm1.54$                             & $19.97\pm2.12$                               & $36.53\pm2.85$   \\
GAIL                            &  $15.04\pm0.67$                             & $17.88\pm0.92$                               & $57.12\pm0.89$   \\
PPO                             &  $13.65\pm0.83$                             & --                                           & $53.48\pm0.76$   \\
SAC                             &  $11.84\pm0.95$                             & --                                           & $43.55\pm4.70$        \\
Ours (policy constraint)        &  $14.72\pm1.13$                             & $19.29\pm2.24$                               & $37.04\pm3.97$   \\ 
Ours (value penalty)            &  $14.26\pm1.24$                             & $19.62\pm2.05$                               & $36.62\pm2.84$   \\\bottomrule
\end{tabular}
}
\end{table}

\begin{figure*}[htp]
    \centering
    \includegraphics[width=\linewidth]{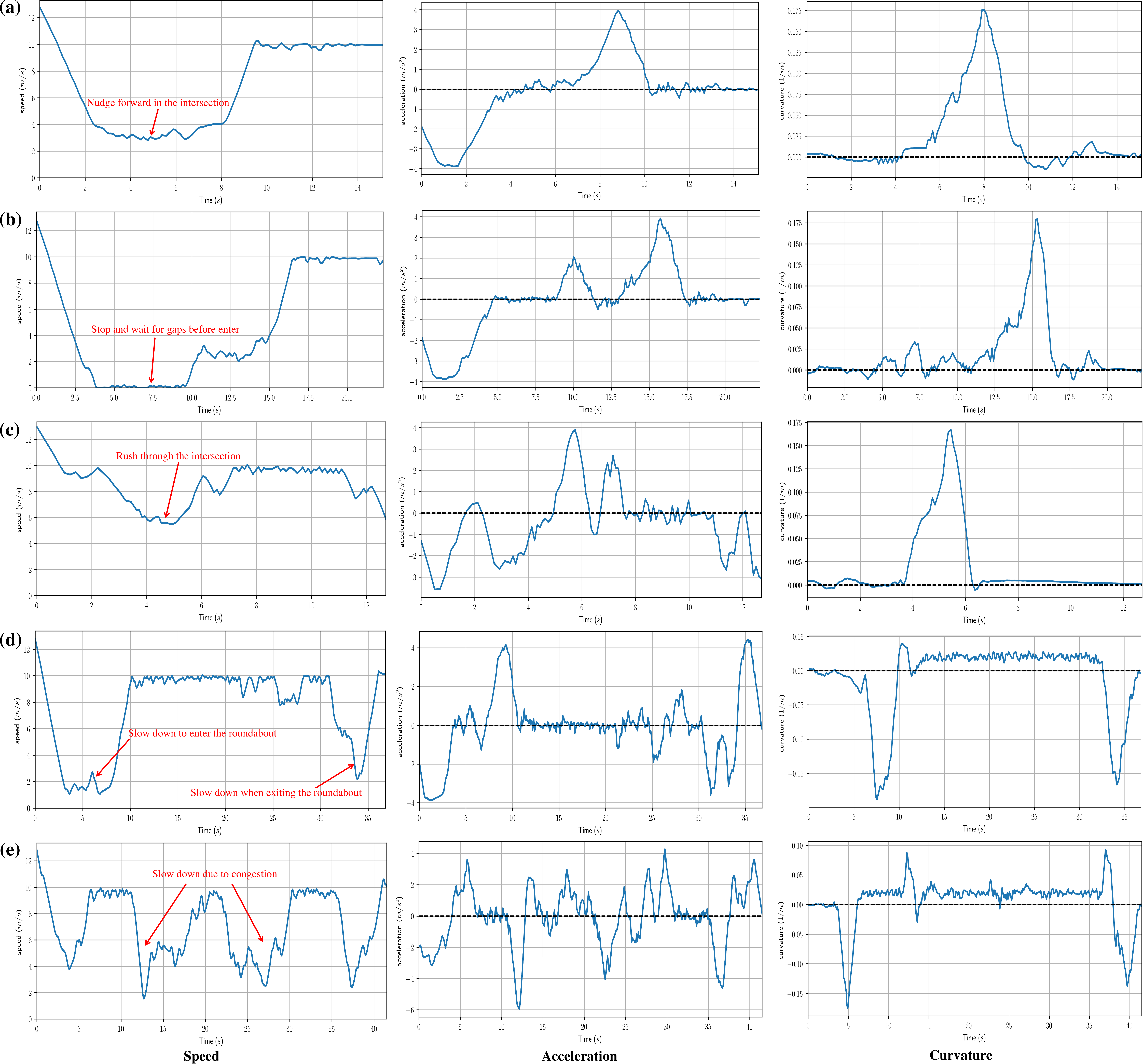}
    \caption{Vehicle dynamic states of different driving policies in the urban driving scenarios: (a) value penalty with aggressive expert policy in the left turn scenario; (b) value penalty with conservative expert policy in the left turn scenario; (c) SAC in the unprotected left turn scenario; (d) value penalty in the roundabout scenario; (e) SAC in the roundabout scenario.}
    \label{fig5}
\end{figure*}

As shown in Fig. \ref{fig5}, we analyze the vehicle dynamic states of different driving policies with some typical cases. Fig. \ref{fig5}(a), (b), and (c) show the speed, acceleration, and curvature of the vehicle in the unprotected left turn scenario for our proposed methods and SAC, respectively. We can notice the different behaviors of aggressive and conservative policies. For aggressive policy, the vehicle will cross the intersection with a low speed (nudging forward); for conservative policy, the vehicle will stop at the intersection, waiting for enough clearance to cross. The stability of the vehicle is guaranteed, as the acceleration and curvature are within a reasonable and comfortable range. For the SAC agent, it will cross the intersection at a higher speed, forcing other vehicles on the main road to yield, and its acceleration profile is more volatile. Fig. \ref{fig5}(d) and (e) display the vehicle dynamic states of our proposed method and SAC in the roundabout scenario. For our proposed method, the vehicle is able to decelerate when entering the roundabout, change to the inner lane to gain a higher speed, and slow down to wait for other vehicles when exiting the roundabout. The acceleration and curvature are smooth and within a comfortable range. However, for the SAC agent, since it only stays in the outer lane, it has to frequently change its speed to yield for other vehicles, and the vehicle could decelerate harshly to avoid a collision.

We also provide supplementary videos to show the behaviors of different driving policies in detail. Overall, our proposed method shows human-like driving behaviors, such as nudging forward in the intersection to show its aggressiveness, stopping and waiting for gaps, changing lanes to gain a higher speed, and decelerating to reduce risk. Although the SAC agent can learn to accomplish the driving tasks, it is not very human-like as it would experience reckless moves like directly cutting into the main road without negotiation and sticking to the slow lane. This is mainly because it only learns to avoid collisions, get a higher speed, and reach the target according to the reward function, without the guidance of human prior knowledge. The PPO and GAIL methods are inefficient and perform poorly in the test, as they have merely learned to keep a low speed and cannot stop to avoid a collision.

%Fig. \ref{fig5}(a) shows the speed curve and some representative scenes of the SAC agent in the unprotected left turn scenario. At around 4s, the ego vehicle slows down to enter the intersection and after a short pause, it goes straight into the intersection at around 6s, forcing the upcoming vehicle on the main lane to yield. Then at around 9s, the ego vehicle has to stop to avoid collision with the front passing vehicle while blocking one lane of the main road. After the front vehicle passes by, the ego vehicle immediately accelerates to cross, forcing the following vehicle to yield.  
%Fig. \ref{fig5}(b) illustrates the driving process of the policy trained by the policy penalty method with aggressive expert policy. As we can see from the speed curve and the scene at 8s,  For the policy penalty method with conservative expert policy, as seen in Fig. \ref{fig5}(c),  
%In Fig. \ref{fig5}(e), the agent trained with the policy penalty method has One human-like characteristic is reflected around 9s to 10s when the ego vehicle faces an upcoming vehicle on the main road. It first decelerates to lower the risk of collision, however, when the conflicting vehicle decelerates, the ego vehicle accelerates again to cross the intersection.

\subsection{Effects of imitative expert policy}

\begin{figure*}[htp]
    \centering
    \includegraphics[width=\linewidth]{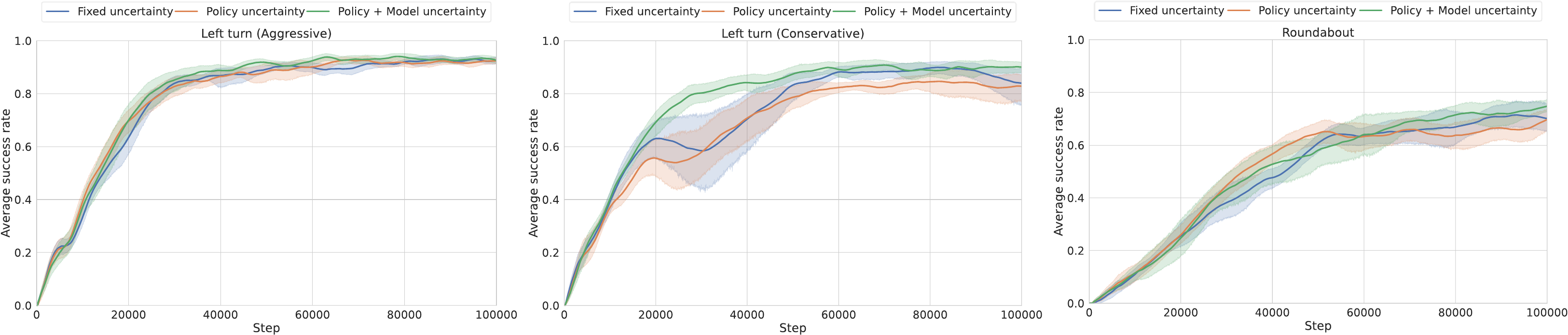}
    \caption{The training processes with different uncertainty estimations for the expert policy: the left shows the average success rate in the unprotected left turn scenario with aggressive expert policy, the middle shows that of the left turn scenario with conservative expert policy, and the right shows that of the roundabout scenario.}
    \label{fig6}
\end{figure*}

\begin{figure*}[htp]
    \centering
    \includegraphics[width=\linewidth]{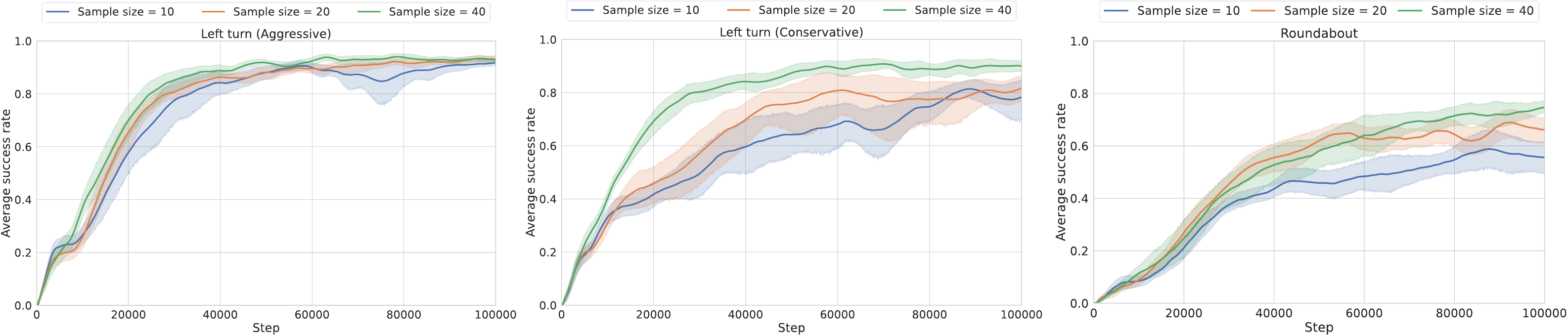}
    \caption{The training processes with different training sample sizes for the expert policy: the left shows the average success rate in the unprotected left turn scenario with aggressive expert policy, the middle shows that of the left turn scenario with conservative expert policy, and the right shows that of the roundabout scenario.}
    \label{fig7}
\end{figure*}

\subsubsection{Uncertainty estimation}
In Section \ref{sec2}, we introduce the ensemble model that can quantify both policy and model uncertainties to obtain the expert policy. Here, we discuss the effects of different uncertainty estimations for expert policy derivation on the training and testing performance of our proposed method. In addition to the proposed ensemble model, we use two other methods to acquire the expert policy. One is a Gaussian distribution policy that only considers policy uncertainty, which means only a single policy network in the ensemble model is used. The other method is a Gaussian distribution policy with fixed uncertainty, which means we take the mean value from a single policy network and set the standard deviation of the distribution as a constant. The fixed standard deviation is set to be 0.2, which is slightly higher than the mean standard deviation given by the ensemble method during training. We employ the value penalty method since it delivers better learning results and the average success rate is employed as the metric.

%One method is vanilla behavioral cloning that outputs a deterministic action value, and we can construct an action distribution, which is assumed to be a Gaussian, with the deterministic action as the mean value and a fixed standard deviation. The other two methods are uncertainty-aware, one of which uses the negative log-likelihood (NLL) to train a Gaussian distributional policy to capture the aleatory uncertainty, and the other uses deep ensemble to obtain a Gaussian mixture distribution to capture both the aleatory and epistemic uncertainties. Besides, we can find that the expert policy with only policy uncertainty estimation does not show a notable benefit over the fixed uncertainty method. 

The training results in Fig. \ref{fig6} indicate that our proposed method would benefit from the expert policy that can estimate both policy and model uncertainty through the ensemble method. This advantage becomes more obvious as the difficulty of the task increases. There is no significant difference between these methods in the unprotected left turn scenario learning aggressive behavior, because the behavior is simple and solution space for this task is limited. However, the ensemble method can stabilize the training process and obtain a higher success rate when learning conservative behavior, while the other two methods could cause an unstable training process and high variances across different trials. This is probably due to that the conservative behavior is more subtle and hard to learn for the expert policy, and a single Gaussian policy network that estimates only the policy uncertainty may not generate a high uncertainty in uncertain states. On the other hand, the ensemble method can emit a larger uncertainty (representing the overall policy and model uncertainty) in uncertain states, which could enhance the exploration of the RL agent. The ensemble method also outperforms the fixed uncertainty method since the fixed uncertainty cannot provide a sensible search range search in different states. In the roundabout scenario, the ensemble method also shows a better result than the other two methods. The data in Table \ref{tab4} about the testing success rate of different expert policy derivation methods also reflects that the ensemble expert policy with both policy and model uncertainty estimations is advantageous, which can provide better testing performance, leading to better robustness and generalization ability. Therefore, the conclusion is that the ensemble method is capable of estimating both policy and model uncertainties and thus better at learning the expert policy (generating expert action distributions), resulting in better performance in guiding the training of RL agents, especially for difficult tasks.

\begin{table}[htp]
\caption{Testing success rate (\%) of the proposed method with different uncertainty estimations for the expert policy (A: aggressive, C: conservative)}
\label{tab4}
\centering

\begin{tabular}{@{}cccc@{}}
\toprule
                            & \tabincell{c}{Unprotected \\ left turn (A)}   & \tabincell{c}{Unprotected \\ left turn (C)}  & Roundabout    \\ \midrule
Fixed uncertainty           & 90                                            & 84                                           &  66           \\
Policy uncertainty          &  \textbf{96}                                  & 90                                           &  72           \\ 
Policy+model uncertainty    &  \textbf{96}                                  & \textbf{96}                                  &  \textbf{84}  \\\bottomrule
\end{tabular}

\end{table}

\subsubsection{Training sample size}
Another factor that could affect the quality of the expert policy and training of the RL agent is the sample size of the expert demonstration. Here, we investigate the influence of the training sample size for the expert policy on the training and testing performance of our proposed method. Specifically, we select 10, 20, and 40 demonstration trajectories to train the expert policy respectively, and use them to guide the training of RL agents. The expert policies are trained with the ensemble method and the RL agents are trained using the value penalty method.

The results in Fig. \ref{fig7} reveal that increasing the training sample size is helpful to train the expert policy and consequently to the training of RL agents. The difference in the sample size becomes more significant with the increase in task complexity. In the unprotected left turn scenario learning the aggressive behavior (relatively simple), a small amount of demonstration trajectories is able to guide the RL training effectively, improving the policy to a satisfactory level, though the learning efficiency and stability are compromised. However, in learning the conservative behavior (relatively hard), a larger training sample size can lead to a better performance in terms of average success rate and learning speed. In the roundabout scenario, the difference between the sample sizes is more notable in the obtained success rate at the end of training. The results in Table \ref{tab5} about the testing success rate of different sample sizes demonstrate that extending training samples is also beneficial for the test performance, especially for more difficult tasks. The possible reason is that increasing training samples with diverse demonstration data enables the expert policy to be more accurate at generating the reference actions and thus providing more accurate action distribution to regularize the RL agent's behavior.

\begin{table}[htp]
\caption{Testing success rate (\%) of the proposed method with different training sample sizes for the expert policy (A: aggressive, C: conservative)}
\label{tab5}
\centering
\begin{tabular}{@{}cccc@{}}
\toprule
                   & \tabincell{c}{Unprotected \\ left turn (A)} & \tabincell{c}{Unprotected \\ left turn (C)}  & Roundabout    \\ \midrule
Sample size = 10   &  92                                         & 70                                           &  60           \\
Sample size = 20   &  94                                         & 80                                           &  78           \\ 
Sample size = 40   &  \textbf{96}                                & \textbf{96}                                  &  \textbf{84}  \\\bottomrule
\end{tabular}
\end{table}
 
\subsection{Discussions}
%Human-like is a critical objective for the autonomous driving industry targeting widespread deployment in the real world. The essence of human-like driving can be summarized as interaction awareness and personalization. These characteristics can be potentially realized through reinforcement learning techniques, because reinforcement learning can learn to interact with other agents on the road without the need of explicitly modeling the environment, and the preferences of different people can be tuned by the reward function. However, the biggest problems of reinforcement learning lie exactly in these two features: one is the low sample efficiency, which requires a massive amount of interactions, and the other is that the reward function of individuals is hard to specify. 

The two biggest problems of RL fall into two categories: one is the low sample efficiency, which requires a massive amount of interactions, and the other is that the reward function is hard to specify. Our proposed method can shed light on how to tackle the issues by introducing human prior knowledge. Specifically, we propose to distill human knowledge, in the form of human demonstration of executing a driving task, into the imitative expert policy, which is subsequently used to guide the training of RL agents. As manifested by the results given in the previous subsections, the strengths of our method are the improved sample efficiency (60\% compared to the state-of-the-art RL algorithm) and ease of requirement of designing reward functions (only using sparse reward), which can hopefully facilitate the application of RL-enabled human-like autonomous driving systems. That being said, some weaknesses of the proposed method should be acknowledged. The major drawback is that we use human experts' high-level decisions as demonstration data, which does not scale to real-world scenarios when only the trajectories are available. Besides, more hyperparameters are introduced in the proposed algorithm and it is time-consuming to tune them to achieve the best performance. Therefore, future work will focus on utilizing naturalistic human driving data to learn expert policies and guide the training of RL agents. Moreover, more methods for deriving the expert policy and generalizing the proposed method to other scenarios are worth being explored.

\section{CONCLUSIONS}
\label{sec6}
In this paper, a framework incorporating human prior knowledge and deep reinforcement learning is proposed and applied in autonomous driving scenarios. Our proposed method consists of three key steps: expert demonstration, policy derivation, and reinforcement learning. First of all, a human expert demonstrates their execution of the task potentially with their own preferences in the expert demonstration step. In the policy derivation step, the expert policy is obtained using imitation learning and uncertainty estimation using expert demonstration data. Finally, in the reinforcement learning step, we propose an actor-critic-based RL algorithm that can incorporate the imitative expert policy into RL training. Specifically, two methods are proposed to regularize the KL divergence between the RL agent's policy and the imitative expert policy, namely value penalty and policy constraint. In order to validate our method, two driving scenarios are designed, and two distinct driving behaviors are demonstrated by the human expert in the left turn scenario. The training results reveal that our method can not only achieve the best performance, but also significantly improve the sample efficiency in comparison with the baseline algorithms. When it comes to testing the learned driving policies in different traffic, the results also verify the outstanding performance of our method with the highest success rate. Besides, the agent is able to perform different driving behaviors given by the human expert, as well as some human-like features. Moreover, we find out that the value penalty method is generally better performing in the environment with sparse reward, and using the ensemble expert policy with both policy and model uncertainty estimations, as well as increasing training examples, can lead to better performance, especially for more difficult tasks. 

%review the main points of the paper, elaborate on the importance of the work or suggest applications and extensions

%\addtolength{\textheight}{-12cm}   % This command serves to balance the column lengths
                                  % on the last page of the document manually. It shortens
                                  % the textheight of the last page by a suitable amount.
                                  % This command does not take effect until the next page
                                  % so it should come on the page before the last. Make
                                  % sure that you do not shorten the textheight too much.

%%%%%%%%%%%%%%%%%%%%%%%%%%%%%%%%%%%%%%%%%%%%%%%%%%%%%%%%%%%%%%%%%%%%%%%%%%%%%%%%
\bibliographystyle{IEEEtran}
\bibliography{IEEEexample}

\end{document}